\documentclass[journal]{IEEEtran}

\usepackage{amsmath}
\usepackage{mathrsfs}
\usepackage{graphicx}
\usepackage[colorlinks=true, allcolors=blue]{hyperref}
\usepackage{titlesec}
\usepackage{amsfonts}
\usepackage{subcaption}
\usepackage{algorithm}
\usepackage{algorithmic}
\usepackage{booktabs}
\usepackage{array}
\usepackage{comment}
\usepackage{multirow}
\usepackage{pifont}

\ifCLASSINFOpdf
\else
\fi

\begin{document}

\title{A Generative Model Enhanced Multi-Agent Reinforcement Learning Method for Electric Vehicle Charging Navigation}

\author{Tianyang~Qi,
        Shibo~Chen,
        and~Jun~Zhang,~\IEEEmembership{Fellow,~IEEE} }
        
%
%
%

\maketitle

\begin{abstract}

With the widespread adoption of electric vehicles (EVs), navigating for EV drivers to select a cost-effective charging station has become an important yet challenging issue due to dynamic traffic conditions, fluctuating electricity prices, and potential competition from other EVs. The state-of-the-art deep reinforcement learning (DRL) algorithms for solving this task still require global information about all EVs at the execution stage, which not only increases communication costs but also raises privacy issues among EV drivers. To overcome these drawbacks, we introduce a novel generative model-enhanced multi-agent DRL algorithm that utilizes only the EV’s local information while achieving performance comparable to these state-of-the-art algorithms. Specifically, the policy network is implemented on the EV side, and a Conditional Variational Autoencoder-Long Short Term Memory (CVAE-LSTM)-based recommendation model is developed to provide recommendation information. Furthermore, a novel future charging competition encoder is designed to effectively compress global information, enhancing training performance. The multi-gradient descent algorithm (MGDA) is also utilized to adaptively balance the weight between the two parts of the training objective, resulting in a more stable training process. Simulations are conducted based on a practical area in Xi’an, China. Experimental results show that our proposed algorithm, which relies on local information, outperforms existing local information-based methods and achieves less than 8\% performance loss compared to global information-based methods.

\end{abstract}

\begin{IEEEkeywords}
Charging navigation, multi-agent deep reinforcement learning, generative model.
\end{IEEEkeywords}

%
\IEEEpeerreviewmaketitle

\section{Introduction}

\subsection{Background and Motivation}

\IEEEPARstart{E}{lectric} Vehicles (EVs) have seen significant advancements and adoption in recent decades \cite{1}. However, due to multiple factors, such as dynamic traffic conditions and fluctuating electricity charging prices, EV drivers often face challenges in selecting optimal EV charging stations (EVCSs) \cite{3} and planning optimal routes \cite{5}. Moreover, when an EV driver has considered all the above factors, there is still a risk that other drivers may make the same decision. This can lead to congestion and substantial waiting time at some EVCSs while leaving other charging stations underutilized. This phenomenon is referred to as the “Future Charging Competition” (FCC) problem \cite{6}, as illustrated in Fig. 1. Consequently, navigating each EV to a suitable EVCS so as to minimize the total cost\footnote{Total cost includes each EV's charging cost and time cost, where time cost consists of driving time on the road and queuing time at an EVCS.} remains an important yet challenging problem \cite{7,8}.  

\begin{figure}[!ht]
\centering
\includegraphics[width=3.5in]{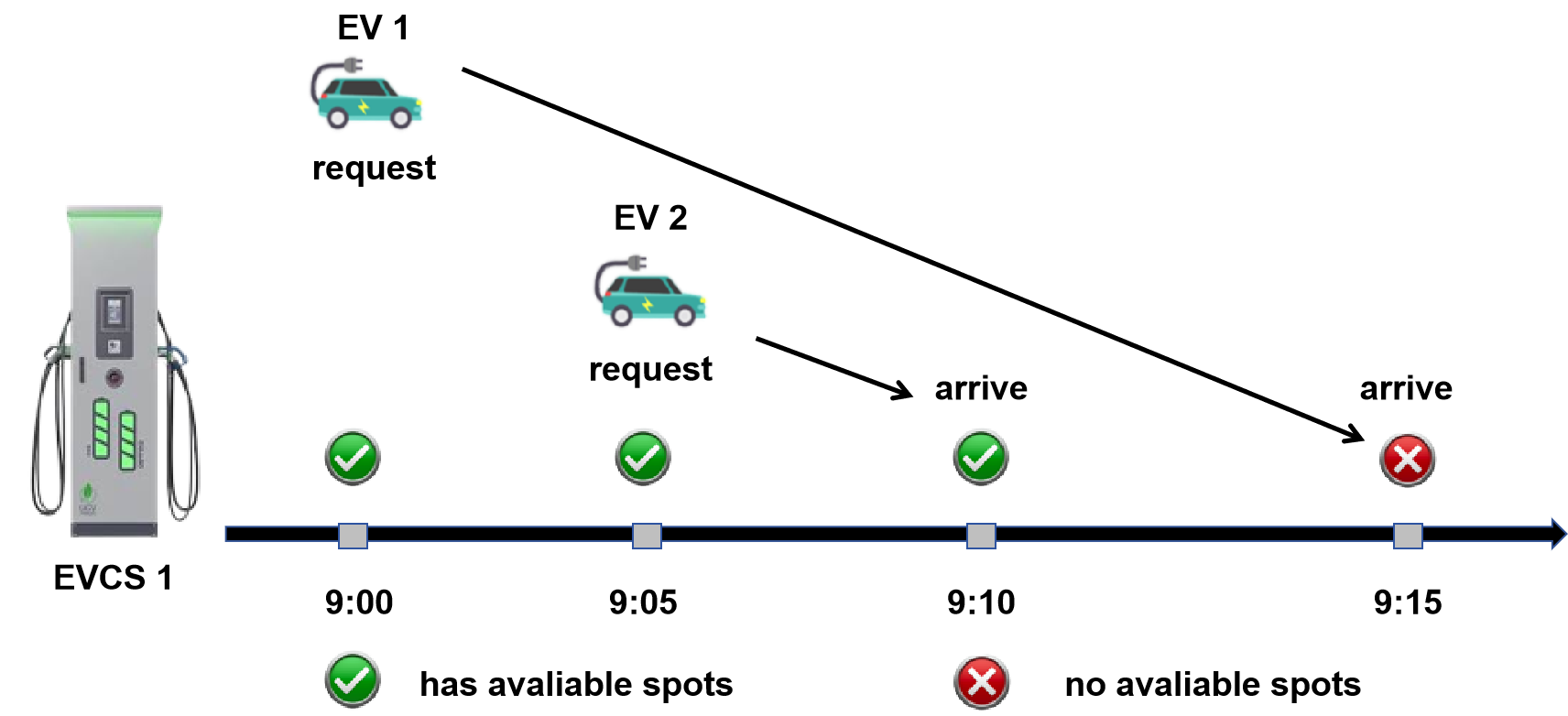}
\captionsetup{font=small}
\caption{An example of the FCC problem: EV 1 decides to charge at 9:00 and arrives at the EVCS at 9:15. However, later at 9:05 another EV 2 also decides to charge, but it arrives earlier at 9:10. This results in EV 1 waiting at EVCS at 9:15. This FCC problem is not known to EV 1 beforehand. But it causes EV 1 substantial inconvenience and should be accounted for when making her charging decisions.}
\label{fig_sim}
\end{figure}

Early works on solving the EV charging navigation problem rely on two types of algorithms: optimization techniques such as linear programming, integer programming, and dynamic programming \cite{10,11,12,13}; heuristic algorithms such as $A^{*}$ and ants algorithm \cite{14,15,16,17}. However, both types   necessitate prior knowledge of all parameters in the whole region, which is difficult to achieve in practice. Under the uncertainty of the system parameters, the optimization algorithms require high computation complexity while the heuristic algorithms lack accuracy. 

Differing from the aforementioned solutions, deep reinforcement learning (DRL) empowers real-time and online decision-making capabilities. However, previous DRL based methods such as \cite{18,19,20} adopted the “Centralized Training Centralized Execution (CTCE)” pattern \cite{21}. In CTCE, all information is aggregated into a single DRL network as a global state during both the training and execution stages. These CTCE methods face challenges as the state dimension expands exponentially when the number of EVs increases. In contrast, introducing the “Decentralized Training Decentralized Execution (DTDE)” pattern reduces dimensionality, where each EV makes decisions based solely on its local information. However, the above-mentioned FCC problem \cite{6} cannot be solved through DTDE pattern.

Graph neural network (GNN) \cite{22} based DRL approaches were introduced in \cite{23,24,25,26} to tackle the above challenges in large-scale navigation tasks. They model the traffic network as a graph and leverage graph attention mechanisms to compress the high-dimensional global state of all EVs into a more compact input. However, there are still notable drawbacks in the current state-of-the-art GNN-based DRL navigation algorithms, as elaborated below:

Firstly, previous GNN-based DRL methods must establish communication channels between the navigation platform and all EVs at each decision-making step to access global information for calculating the graph attention, as shown in Fig 2. This incurs substantial communication cost when the number of EVs is high. Potential communication latency and packet loss can deteriorate the performance of these GNN-based DRL methods. Furthermore, frequent communication may also result in the privacy issues.

\begin{figure}[!ht]
\centering
\includegraphics[width=3in]{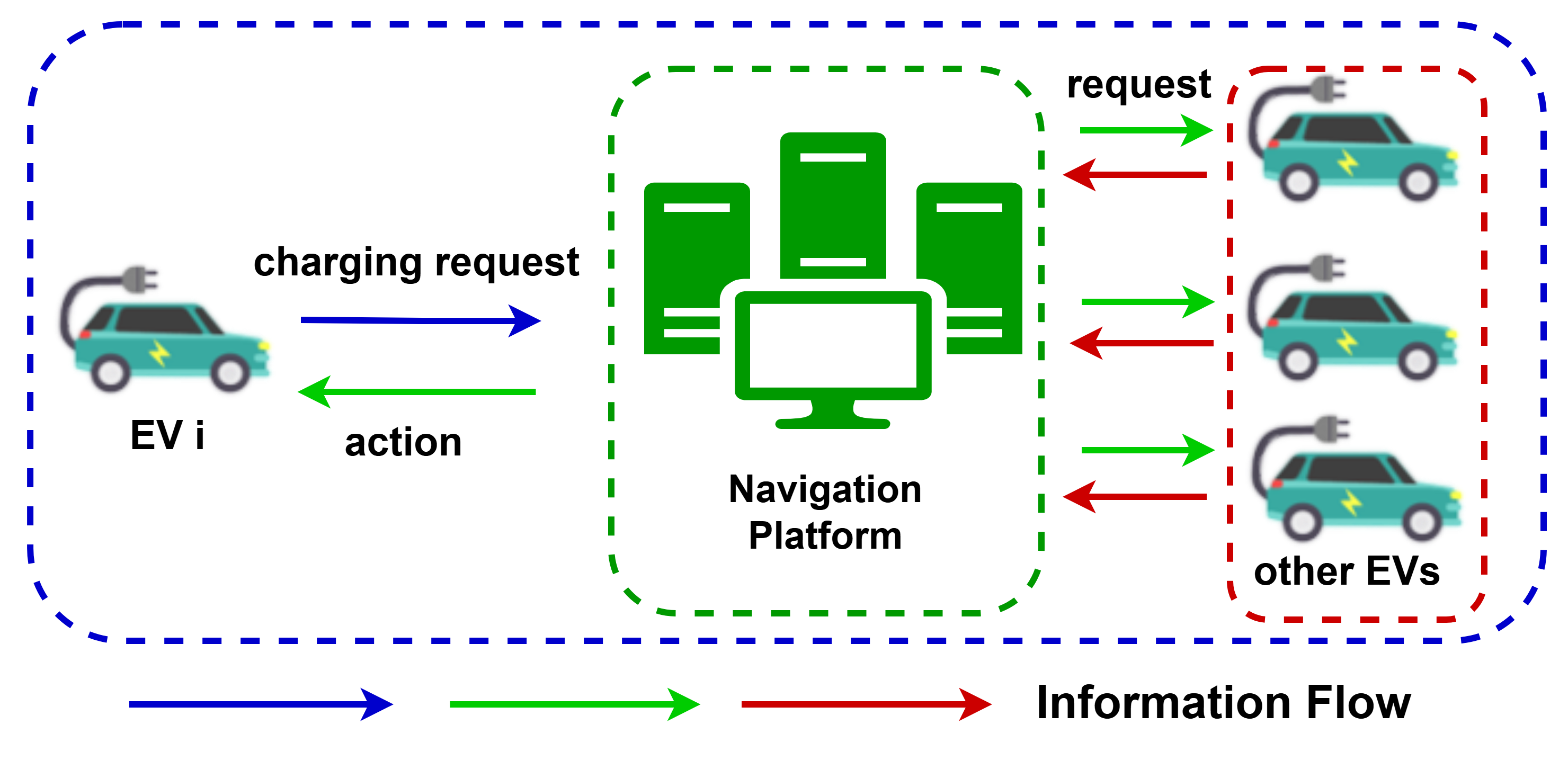}
\captionsetup{font=small}
\caption{The information flow of GNN-based DRL methods during the execution stage. These algorithms still require access to global information at each decision step to calculate the graph attention.}
\label{fig_sim}
\end{figure}

Secondly, previous works assume that the navigation platform directly assigns specific EVCS as the destination of the EV drivers at each decision step, as shown in Fig. 2. However, in practice, a platform should offer “\textbf{advisory}” information to the EV drivers rather than directly issue “\textbf{commands}” when the EV drivers require navigation assistance. 

Thirdly, previous GNN-based DRL methods like \cite{24,26} have structured the data according to the graph of traffic or power network. This formulation requires the EVs to be precisely located on the nodes of the graph at each decision step. However, in practice, some EVs can be positioned on the edges rather than the nodes. In these scenarios the graph attention calculated using the GNN-based methods lacks precision.

\subsection{Literature Review}

\subsubsection{Optimization based Charging Navigation Algorithm}

Optimization based navigation algorithms utilize optimization techniques to determine cost-minimum routes \cite{10,11,12,13,34}. A hierarchical game approach \cite{10} is proposed for EV path planning. At the upper level, a non-cooperative game is proposed to model the competition between EVCSs, while multiple evolutionary games are formulated at the lower level for EVs to choose a suitable EVCS as the destination. In \cite{11}, a simplified charge-control (SCC) based programming algorithm is presented for navigation which can simplify the charging control decisions within an SCC set. \cite{12} proposes an evolutionary game model for navigation in a complex network which considers the social relationships and mutual learning among users. In \cite{13}, a mixed-integer linear programming model is developed to tackle the traveling salesman problem that takes into account the impact of time-of-use electricity pricing, predefining distribution routes and specifying charging points along the route for electric logistics vehicles.  \cite{34} develops a bi-layer navigation model which coordinates both the transportation network and distribution network. The model minimizes the total navigation cost through the upper level and reduces energy exchange among EVCSs and power grid upon the lower level.

\subsubsection{Heuristic Charging Navigation Algorithm}
Heuristic charging navigation algorithms utilize the heuristic rules or strategies \cite{14} to guide the decision-making process, such as prioritizing charging at EVCSs, minimizing detours for charging, or considering traffic conditions \cite{15,16,17}. In \cite{15}, a driving strategy based on the distributed ant system algorithm was designed for EV charging navigation. \cite{16} presents a comprehensive framework that considers different problem variants, speedup techniques, and develops three solution algorithms: an exact labeling algorithm, a heuristic labeling algorithm, and a roll-out algorithm. In \cite{17}, a cooperative-A* algorithm was proposed to solve the cooperative real-time EV planning problem.

\subsubsection{DRL based Charging Navigation Algorithm}
 DRL acquires decision-making strategies through interacting with environment \cite{18,19,20,23,24,25,26}. The most common-used DRL framework for EV charging navigation is Deep-Q-Network (DQN) \cite{27}. In \cite{19}, the EV charging navigation problem is firstly framed as a Markov Decision Process (MDP), and DQN is developed aiming at capturing the unique state of the unknown environment, ultimately providing the optimal travel route and EVCS selection policy for the EV drivers. In another study \cite{20}, a Constrained MDP was formulated to design a constrained charging/discharging scheduling strategy to minimize charging costs while ensuring that EVs achieve a full charge. 
 
 For GNN-based DRL algorithms, \cite{23} uses a graph convolutional network to extract the environment information from the coupled power-traffic system of the urban area. In \cite{24}, a bi-level graph based DRL method is proposed, where the upper level focuses on selecting the optimal EVCS, while the lower level is dedicated to routing EVs efficiently. In \cite{25}, a bi-timescale GNN based DRL algorithm was introduced, of which at the slow timescale the algorithm focuses on resolving the distribution locational marginal pricing of the node and employs multi-agent DRL to address real-time EV requests on the fast timescale. In \cite{26}, three supplementary models concerning the traffic network, charging station, and EV driver are incorporated into the GNN-based DRL framework to enhance the overall performance of the algorithm.

\subsection{Paper Contributions}

To tackle the aforementioned drawbacks, this paper proposes a novel generative recommendation model enhanced DRL framework for the real-time EV charging navigation problem. Our approach focuses on providing the recommendation information to EV drivers based solely on their \textbf{local information}. To the best of the authors' knowledge, this paper is the first to utilize a generative model to represent the distribution of the global state, achieving comparable performance against the state-of-the-art DRL methods but without \textbf{requiring access to global information} during the execution stage. Our proposed method has the potential to significantly reduce the \textbf{communication cost}, improves the \textbf{scalability} of the navigation algorithm and the \textbf{privacy} of EV drivers. A detailed comparison with previous methods is shown in Table I. The contributions include the following aspects:  

\begin{table}[h!]
\centering
\captionsetup{font=small}
\caption{Comparation of methods for charging navigation.}
\begin{tabular}{@{}ccccc@{}}
\toprule
\textbf{Type} & \textbf{method} & \textbf{model-free} & \textbf{FCC-aware}& \textbf{local info}  \\ 
\midrule
\multirow{5}{*}{Optimization} & \cite{10}         & \text{\large\ding{55}} & \text{\large\checkmark} & \text{\large\ding{55}} \\ 
& \cite{11}        & \text{\large\ding{55}} & \text{\large\checkmark} & \text{\large\ding{55}} \\ 
& \cite{12}         & \text{\large\ding{55}} & \text{\large\checkmark} & \text{\large\ding{55}} \\ 
                     & \cite{13}    & \text{\large\ding{55}} & \text{\large\checkmark} & \text{\large\ding{55}} \\      
                     & \cite{34}    & \text{\large\ding{55}} & \text{\large\checkmark} & \text{\large\ding{55}} \\ 
\midrule
\multirow{3}{*}{Heuristic}  
                         & \cite{15}         & \text{\large\checkmark} & \text{\large\checkmark} & \text{\large\ding{55}} \\
                         & \cite{16}         & \text{\large\checkmark} & \text{\large\checkmark} & \text{\large\ding{55}} \\
                         & \cite{17}         & \text{\large\checkmark} & \text{\large\checkmark} & \text{\large\ding{55}} \\

\midrule
\multirow{5}{*}{DRL} & \cite{19}      & \text{\large\checkmark} & \text{\large\ding{55}} & \text{\large\checkmark} \\
                         & \cite{24}        & \text{\large\checkmark} & \text{\large\checkmark} & \text{\large\ding{55}} \\ 
                         & \cite{25}        & \text{\large\checkmark} & \text{\large\checkmark} & \text{\large\ding{55}} \\
                         & \cite{26}       & \text{\large\checkmark} & \text{\large\checkmark} & \text{\large\ding{55}} \\
                        & \textbf{Ours}        & \text{\large\checkmark} & \text{\large\checkmark} & \text{\large\checkmark} \\ 
\bottomrule
\end{tabular}
\end{table}

Firstly, we introduce a novel charging navigation algorithm that integrates a generative model with a sequential charging request encoder. This algorithm utilizes only local information from each individual decision-making EV while achieving performance similar to that of algorithms that have access to global information.

Secondly, we observe that the distribution of the global states is hard to represent due to the dynamic number of EVs and high dimensions of the data when implementing the generative model. To address this, we present a novel FCC-based encoder which compresses the dynamic global state into a fixed-dimensional tensor. Its dimension is always equal to the number of EVCS in the region, regardless of how large and dynamic the number of EVs is considered.  

Finally, due to the differing sensitivities of the DRL model and generative model to the training loss, simply summing up both losses together can lead to significant performance degradation. Thus, we introduce the Multi-Gradient Descent Algorithm (MGDA) into our DRL framework to adaptively balance the update steps between the DRL model and generative model to ensure stable training.

The rest of this article is organized as follows. The problem statement and system modeling are introduced in Section II. Then, our proposed solution is presented in Section III. Case studies are reported in Section IV to verify our proposed methodology. Finally, Section V concludes the article.

\section{Problem Statement and System Modeling}

\subsection{Formulation of EV Charging Navigation}

The objective of the EV charging navigation problem is to minimize the total cost of all EVs that require charging in the region. The cost consists of the battery energy consumption cost on the road $C_{road}$, the charging cost $C_{ch}$ at the EVCS, as well as the total time spent including the driving time $T_{road}$ on road and waiting time $T_{wait}$ at EVCS. 

An EV charging navigation task starts when all EVs are at their initial position and ends when all EVs reach suitable EVCSs. The traffic network in the whole region is modeled as a graph. In the traffic graph, the roads are formulated as the edges and the crossroads as nodes, as shown in Fig 3. Without loss of generality, we suppose that all EVCSs are located on the nodes of the graph.  

\begin{figure}[!ht]
\centering
\includegraphics[width=3.5in]{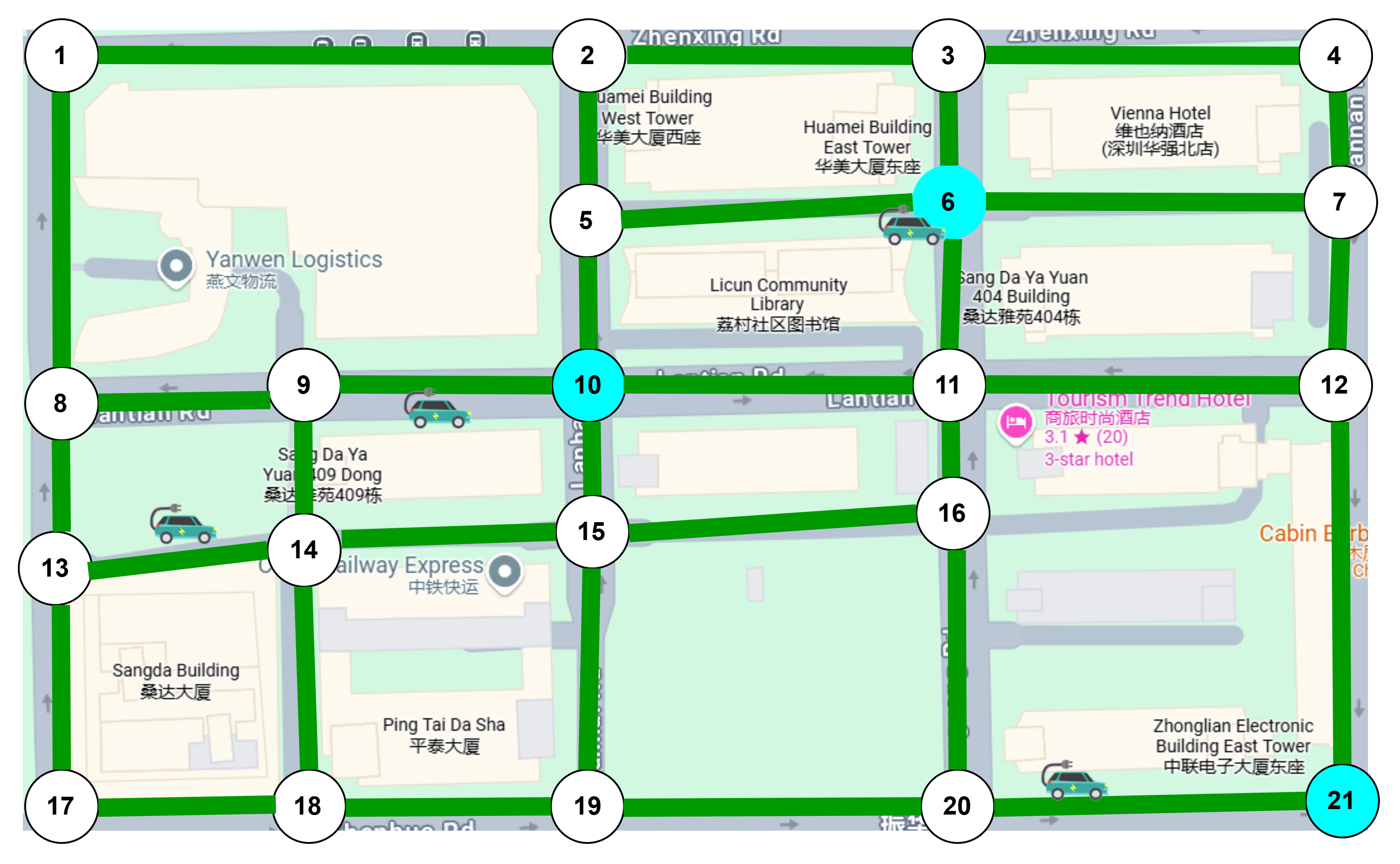}
\captionsetup{font=small}
\caption{An example of the traffic network in Shenzhen, China. The blue nodes are the locations of the EVCSs.}
\label{fig_sim}
\end{figure}

\subsection{Formulation of Dec-POMDP}
Considering the problem's nature of sequential decision making and the limited vision of each EV, we formulate the problem as a finite decentralized partially observable Markov decision process (Dec-POMDP). The decision step occurs as some EVs reach a node, while other EVs are still driving along the edges of the traffic network. Thus, in each decision step, we only focus on those EVs that reach the nodes and consider their decision-making.

We assume that there are $M$ nodes in the traffic graph. Meanwhile, there are $N$ EVs and $K$ EVCSs in the region. At decision step $t$, the global state consists of all EV agents' locations $[p^{n}_{t}]$, current state-of-charge (SOC) $[SOC^{n}_{t}], n \in N$, each road's average  velocity $[v^t_{ij}],(i,j) \in M$ in the traffic graph and each EVCS $k$'s charging price $[\lambda_k^{t}], k \in K$. A decision-making EV agent $n$'s partial observation $o_{t}^n$ includes its location $p_{t}^n$ and current SOC $SOC_{t}^n$, a piece of recommendation information (RI) ${RI}_{t}^n$ received from the recommendation platform. The details of RI will be elaborated later. EV agent $n$ executes an action by selecting an EVCS $k$ in the traffic network as the destination. Then based on the current traffic flow and charging prices, EV agent $n$ determines a route $L_{k}^{n}$ following the Dijkstra shortest-path algorithm. A reward $r_{t}^n$ is observed until EV agent $n$ reaches a next node following $L_{k}^{n}$ at a later time step $(t+1)_{n}$. At that time, EV agent $n$ observes the new $o_{t+1}^n$, takes action $a_{t+1}^n$, and will later observe the reward $r_{t+1}^n$. Based on the above procedure, the formulation of the Dec-POMDP is defined as follows:

\textit{State} $S_{t}$: The global state at decision step $t$ includes all the aforementioned information in the region, where $n \in N, k \in K, (i,j) \in M$: 

\begin{align}
S_{t} = \left( \{p^{n}_{t}\}, \{SOC^{n}_{t}\}, \{v^{t}_{ij}\},\{\lambda_k^{t}\} \right)    
\end{align}

\textit{Observation} $o_{t}^n$: The observation of an EV agent $n$ at decision step $t$ consists of the following states: 

\begin{equation}
o_{t}^n = \left( p_{t}^n, SOC_{t}^n, RI_{t}^n \right)
\end{equation}

\textit{Action} $a_{t}^n$: Given the observation $o_{t}^n$, EV agent $n$ selects an action, which represents the target EVCS $k$ for charging and the corresponding planned route $L_{k}^{n}$ dictated by the Dijkstra shortest-path algorithm:

\begin{equation}
a_{t}^n =k , \quad k \in K.
\end{equation}

\textit{Transition Function}: State transition occurs when a certain EV agent arrives at one node. For other EV agents that still driving along the edge during that time period, we assume they are moving in the fixed average speed $v_{edge}^t$. The transition is determined by the next step's decision-making EV agent $q$, and the stochastic data including average road velocity and each EVCS's charging price. The transition function of each parameter is shown as follows: 

\begin{subequations} 
\[
\left\{
\begin{aligned}
p^q_{t+1} &= L_k^{q,\text{next node}} && \text{(4a)} \\
p^{\text{other}}_{t+1} &= p^{\text{other}}_{t} + t \cdot v_{\text{edge}}^t && \text{(4b)} \\
SOC^n_{t+1} &= SOC_{t}^n - e^{n,\text{cost}}_t, \quad \text{all } n \in N && \text{(4c)} \\
S_{t+1} &= \left( \{p_{t+1}\}, \{SOC_{t+1}\}, \{v^{t+1}_{ij}\}, \{\lambda^{t+1}\} \right) && \text{(4d)}
\end{aligned}
\right.
\]
\end{subequations}

\noindent where 4(a) and 4(b) denote the position transitions of next step's decision-making EV agent $q$ and other EV agents. 4(c) represents the transition of each EV agent's SOC, where $e^{n,\text{cost}}_t$ represents the energy consumption of EV agent $n$ within the time period between $t$ and $t+1$. 4(d) records the transition of the global state.

\textit{Reward} $r^n_{t}$: For a certain EV agent $n$, only until it reaches the next node on $L_{k}^{n}$ will it receive its reward $r^n_{t}$, as shown below:

\begin{equation}
r^n_t = 
\begin{cases} 
-\alpha\lambda_k^{t+1} d_{ij} - \pi d_{ij}/v^{t}_{ij}, & \text{next node} \neq L^{\text{end}} \\ 
-e^{n,cost}_t \lambda_k^{t+1} - \pi t_k^{\text{wait}}, & \text{next node} = L^{\text{end}}
\end{cases} 
\end{equation}

\noindent where the variable $\alpha$ represents the electricity consumption per kilometer, $\pi$ represents the money cost per minute, $t_{k}^{wait}$ represents the actual waiting time at the EVCS $k$, and $L^{end}$ represents the final node on the planned route $L^n_{k}$, respectively. 

In equation (5), the reward is computed differently based on the EV's location. When the EV is on the road, the reward is determined by the actual battery consumption and driving time between nodes $i$ and $j$. If the EV arrives at an EVCS, the reward is calculated according to the actual charging cost and waiting time cost. We assumed that the charging power at each EVCS remains constant and uniform in a certain time period \cite{19}, implying that the charging time is proportional to the charging cost.

\textit{Action-Value Function} $Q_n^\psi(o, a)$: The consequence of the EV agent $n$ taking the action $a$ under the observation $o$ and subsequently adhering to the policy $\psi$ is evaluated as the anticipated discounted cumulative reward from time step $0$ to its end step $T$, as illustrated in equation (6):

\begin{equation}
Q_n^\psi(o, a) = \mathbb{E}^\psi \left[ \sum_{t=0}^{T} \gamma^t r^n_{t} \mid o^n_t = o, a^n_t = a \right]
\end{equation}

Here, $Q_n^\psi(o, a)$ represents the action-value function \cite{27}, and $\gamma$ is the discount factor, delineating the equilibrium between immediate rewards and long-term gains. 

\section{Methodology}

The general framework of how our designed navigation algorithm works in each step is shown in Fig. 4. In light of the aforementioned drawbacks of previous methods, we shift the policy network onto the EV side. The recommendation platform only provides advisory information as part of input to EV's policy network rather than directly issuing the action. 

\begin{figure}[!ht]
\centering
\includegraphics[width=3.5in]{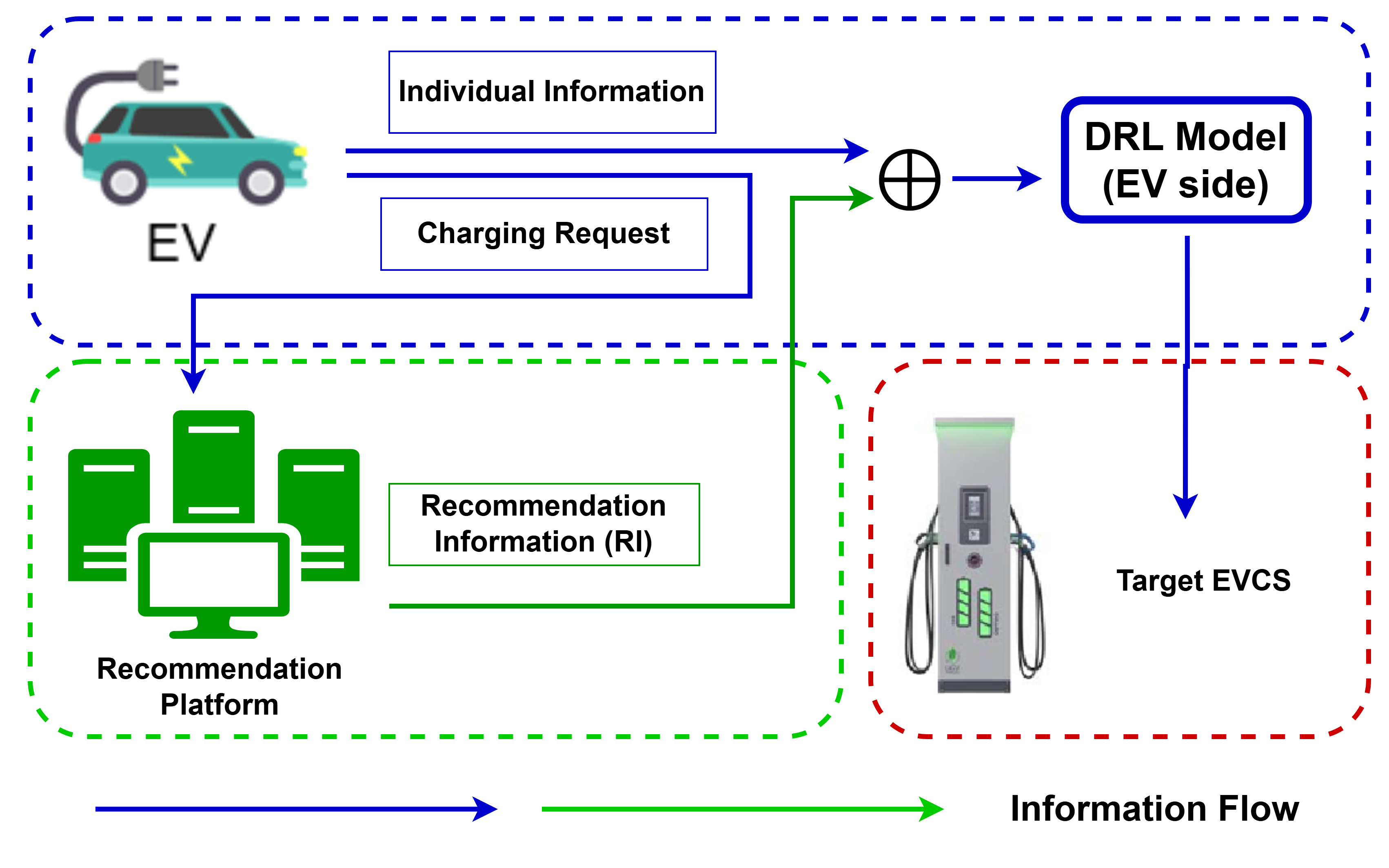}
\captionsetup{font=small}
\caption{Procedures in a decision step in EV charging navigation task.}
\label{fig_sim}
\end{figure}

The procedures of how our framework works are as follows: (1) When an EV agent approaches a node, it sends a charging request, including its current location and SOC to the recommendation platform; (2) Subsequently, the recommendation platform combines this charging request with current road velocity and electric price in the region to generate a piece of RI, which reflects the FCC status of each EVCS at present; (3) The RI is then transmitted back to the EV and aggregated with EV's local observation as input into the DRL policy network which is deployed on the EV side; (4) The DRL policy network finally outputs the next action which determines the target EVCS for charging. Note that in our proposed framework, the recommendation platform can provide advice solely based on current EV's local observation, which mitigates the needs of accessing the global information as well as the need of graph structure of the input data.

Thus, the DRL network and the recommendation model are the two pillars of our charging navigation algorithm. Each will be elaborated upon in the following subsections.

\subsection{DQN Policy Network for Individual EV}
In our method, we utilize the DQN \cite{27} algorithm as the policy network for each EV's decision making. Algorithm 1 highlights the training process of the DQN algorithm for each EV in the charging navigation task. 

We implement a “Centralized Training Decentralized Execution” (CTDE) framework to train the DQN model. Firstly, the DQN parameters $\theta$ are randomly initialized. A target network, mirroring the structure of the DQN, copies its parameters $\hat{\theta}$ from $\theta$. Within the inner \textit{while} loop, commencing from step 8, the DQN model receives RI from the platform and combines it with this EV's local data as input. It then selects an EVCS as the temporary destination based on the $\epsilon$-greedy strategy as an action. Following this, the EV agent progresses using the Dijkstra shortest-path algorithm, observes the reward $r_{t}$ according to equation (5), acquires new data and generates a new observation $o_{t+1}$. The tuple $(o_{t}, a_{t}, r_{t}, o_{t+1})$ is stored in the replay buffer, from which a batch of tuples is extracted to compute gradients for updating the DQN parameters in steps 12-14. Finally, in step 15, the parameters of the target network are synchronized with those of the DQN.

\begin{algorithm}[!ht]
\caption{Training Process of DQN on the EV side}
\begin{algorithmic}[1]
\STATE Randomly initialize DQN parameters $\theta$.
\STATE Initialize target network parameters $\hat{\theta} \gets \theta$.
\FOR{Epoch = 1:1000}
    \FOR{all EVs}
        \STATE Generate the initial state $s_0$.
    \ENDFOR
    \WHILE{not all EVs in the EVCS}
        \STATE Find the decision making EV agent and receive RI. Then the decision-making EV selects an EVCS and calculate the corresponding route via action $a_t$ based on $\varepsilon$-greedy strategy.
        \STATE Take a step, observe the reward $r_t$ and the next observation $o_{t+1}$.
        \STATE Store the tuple $(o_t, a_t, r_t, o_{t+1})$ in replay buffer $\Xi$.
        \STATE Sample a batch $\Phi = \{(o_t, a_t, r_t, o_{t+1})\}$ from $\Xi$.
        \STATE $q_t \gets r_t + \gamma \max_{a \in A} Q(o_{t+1}, a | \hat{\theta})$
        \STATE Calculate the loss function: \\
        $L(\theta_t) = (q_t - Q(o_t, a_t | \theta))^2$
        \STATE Update DQN parameters $\theta \gets \theta - \alpha \nabla L(\theta)$
        \IF{Every $B$ steps}
            \STATE $\hat{\theta} \gets \theta$
        \ENDIF
    \ENDWHILE
\ENDFOR
\end{algorithmic}
\end{algorithm}

\subsection{Generative Recommendation Platform}

To enable appropriate decision-making based solely on the EV agent’s local data while achieving performance comparable to algorithms that rely on global information, we introduce a novel CVAE-LSTM-based generative recommendation platform to provide RI to the EV agent’s policy network. The process of training and executing this recommendation platform is illustrated in Fig. 5.

\begin{figure}[!ht]
\centering
\includegraphics[width=3.5in]{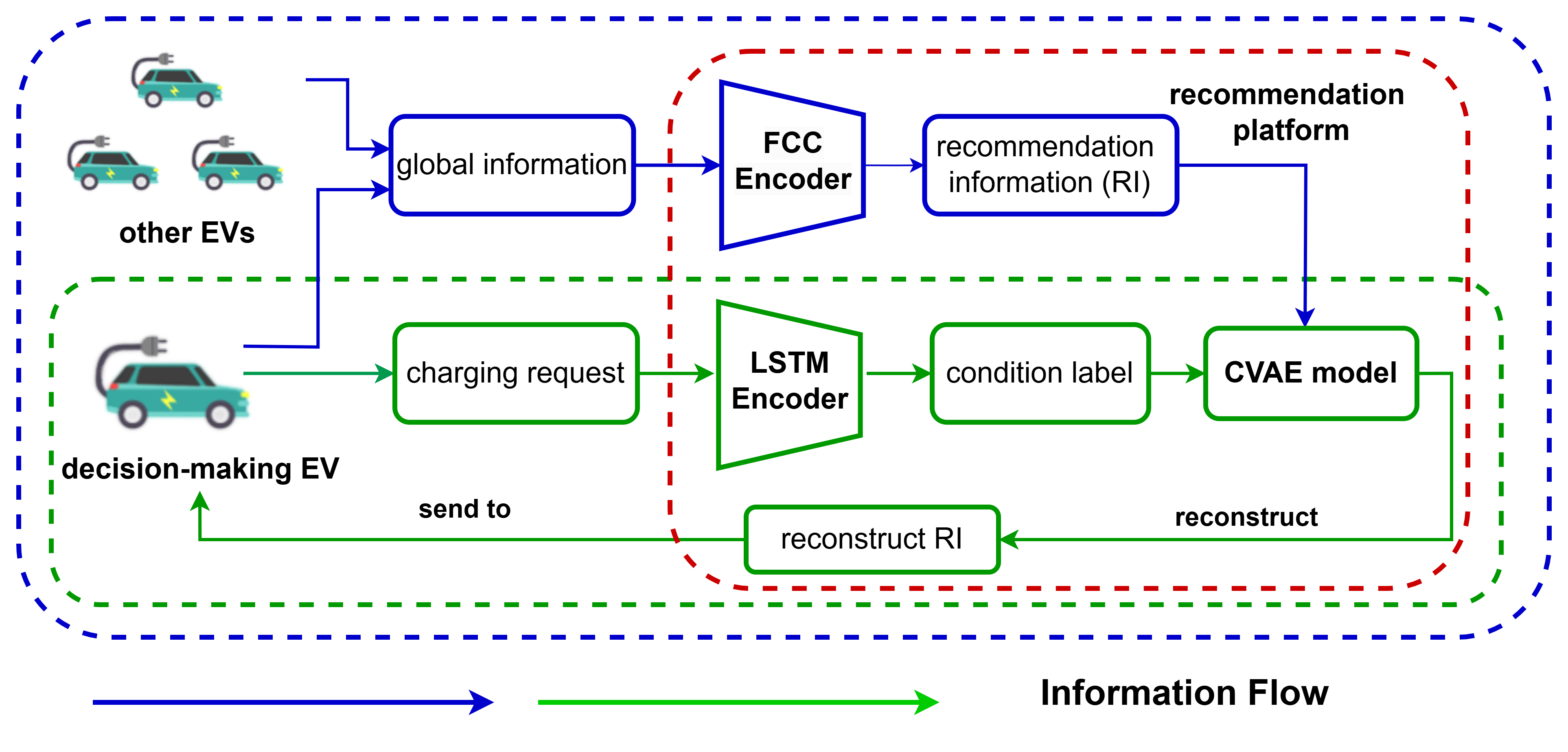}
\captionsetup{font=small}
\caption{The structure of our proposed CVAE-LSTM based recommendation platform. The blue and green lines indicate the information flow, with the blue lines only exist in the training stage while the green lines exist in both the training and execution stage.}
\label{fig_sim}
\end{figure}

The procedures of our proposed framework are as follows: 

First, during the training stage, we assume that both the EV agents and the recommendation platform have access to global information, as required by the CTDE pattern. We then employ a novel FCC-based encoder to compress the varying-dimensional global state into a fixed-dimensional FCC tensor, the dimension of which is solely determined by the number of EVCSs. Subsequently, this FCC tensor serves as the input to a Conditional Variational Autoencoder (CVAE) model. This FCC tensor is considered as RI, which will be reconstructed and sent to the EV agent during the execution stage.

Second, during the execution stage, when an EV requires RI for decision-making, it first sends a charging request to the recommendation platform, where it subsequently receives and maintains a sequential queue of charging requests over several steps. We utilize a Long Short-Term Memory (LSTM) encoder to capture the time-sequential nature of this data. Once encoded by the LSTM, the data is used as the condition label for the CVAE model. Based on this condition label, the CVAE model reconstructs the RI corresponding to that step’s original global state.  

The details of each part of the recommendation model are introduced in the following subsections:

\subsubsection{CVAE-LSTM model of Recommendation Framework}

The objective of the recommendation platform is to provide effective RI to EV agents based solely on their local information, meanwhile achieving comparable performance against methods that rely on global data. Due to the ability to model conditional distributions, CVAE model \cite{28} is well-suited for the scenarios where the relationship from input to output is not strictly one-to-one, but rather one-to-many. We therefore employ CVAE model to learn the distribution of the global state and compress all the steps' global states into a latent space at the training stage, aiming at later reconstructing them at the execution stage with only local observation. The condition label in the CVAE model refers to which one of the global states should be reconstructed from the compressed latent space.

\begin{figure}[!ht]
\centering
\includegraphics[width=3.5in]{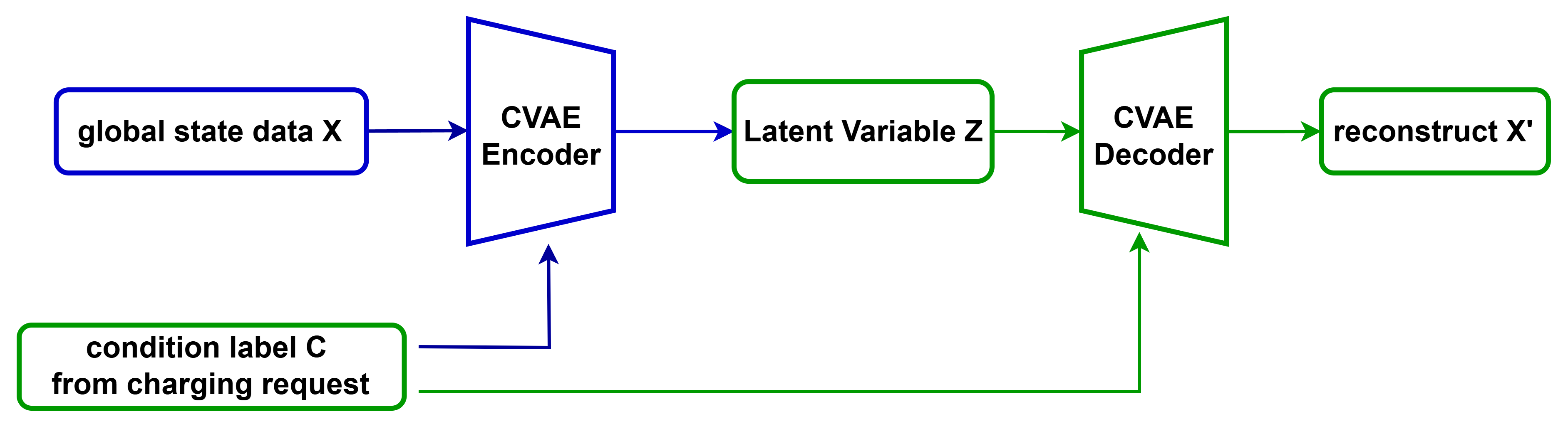}
\captionsetup{font=small}
\caption{The framework of the CVAE model in our recommendation platform. The blue lines only exist in the training stage and the green lines exist in both stages.}
\label{fig_sim}
\end{figure}

The framework of the CVAE model in our algorithm is shown in Fig. 6. At the training stage, the CVAE model first concatenates the global data $x$ which indicates each step's $S_{t}$, and its related condition label $c$ from the charging request data. Then the CVAE encoder $q_\phi(\mathbf{z}|\mathbf{x}, \mathbf{c})$ compresses them into a latent variable z. Next, based on the condition label $c$ and the latent variable $z$, the CVAE decoder $p_\theta(\mathbf{x'} | \mathbf{c}, \mathbf{z})$ reconstructs $x$ as $x'$. The objective of the CVAE model is to maximize the  variational lower bound, which is shown in equation (7):

\begin{align*}
\tilde{\mathcal{L}}_{\text{CVAE}}(\mathbf{x}, \mathbf{x'}; \theta, \phi) = 
- KL\left(q_\phi(\mathbf{z} \mid \mathbf{x}, \mathbf{c}) \parallel p_\theta(\mathbf{z} \mid \mathbf{x}, \mathbf{c})\right) \\
+ \frac{1}{L} \sum_{l=1}^{L} \log p_\theta(\mathbf{x'} \mid \mathbf{z}, \mathbf{c}) \tag{7}
\end{align*}

\noindent where $KL$ denotes the Kullback–Leibler (KL) divergence. This divergence measures how much information is lost when using $q$ to represent the prior on $z$.  The second item is to measure the expected logarithmic probability error of the reconstruction. A larger error indicates that the decoder is unable to reconstruct the data precisely.

During the execution stage, CVAE model reconstructs the data solely based on the condition label $c$ and the latent variable $z$, utilizing them as input to the decoder $p_\theta(\mathbf{x'} | \mathbf{c}, \mathbf{z})$ to generate the data we need. 

To obtain a proper condition label, we observe that the recommendation platform receives a charging request at each decision step. Over several steps, the recommendation platform will maintain a queue of charging request data in chronological order. At each decision step, a one-to-one relationship exists between the time-ordered queue of data and that step's global state. However, directly utilizing this queue of data as the condition label is not practical. It is costly for the recommendation platform to maintain such a long queue as the number of the steps increases. On the other hand, LSTM model can incorporate temporal correlations between sample sequences to address the partial observability of the environment, while managing time-ordered data without requiring extensive storage. Thus we encode this time-ordered queue of data using an LSTM encoder. In this way, in each step the recommendation platform only needs to input one charging request data into the LSTM model, as illustrated in Fig. 7. 

\begin{figure}[!ht]
\centering
\includegraphics[width=3.5in]{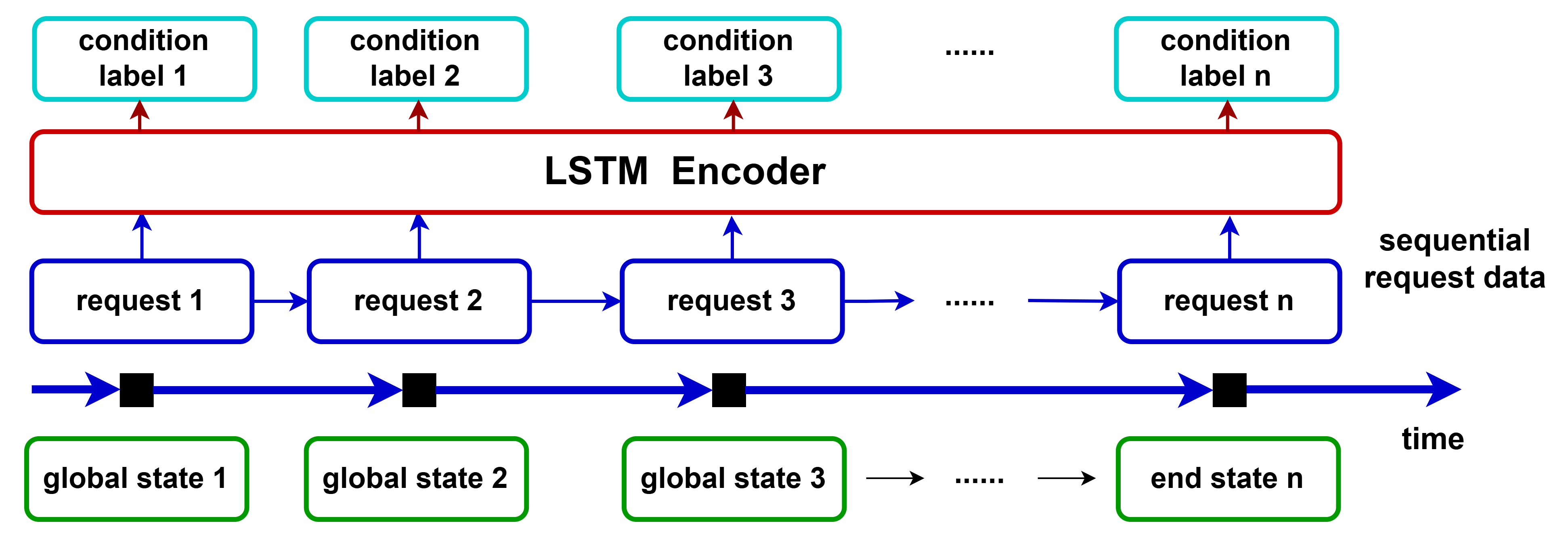}
\captionsetup{font=small}
\caption{LSTM encoder to provide condition label for the CVAE model. By utilizing LSTM, the queue of data is recorded in a single model model under partial observation without requiring extensive storage resource.}
\label{fig_sim}
\end{figure}

 Finally at the execution stage, the latent variable $z$ and the condition label from LSTM encoder are used by the recommendation platform to provide RI. Through our CVAE-LSTM framework, the RI corresponding to the current global state is effectively generated and sent to the EV agent for decision-making at each step without accessing the global state.

\subsubsection{FCC-based Information Encoder}

In subsection 1), we introduce a novel CVAE-LSTM based recommendation platform to offer RI solely based on the EV agent's local information. However, a significant challenge arises when implementing this framework directly in practical tasks: The number of EVs within a given region is constantly fluctuating and the global states are often high dimensional. For the CVAE encoder, directly utilizing such global data as inputs to compress is exceptionally challenging.

As depicted in Fig. 1, the rationale behind making decisions based on the global information is to mitigate the impact of potential FCC among EVs, thereby reducing the queuing time cost. This inspires us to compress the global states by estimating the “\textbf{expected queuing time}” when the EV agent arrives at the EVCS in the future. By doing so, we can compress the original global state into a fixed-dimensional tensor, which we call the “\textbf{FCC tensor}”, whereby the dimensions are associated solely with the number of EVCSs in the region. Each element in this tensor signifies the anticipated future queuing time if the EV agent opts for that EVCS as its destination at the current decision step. 

We aim to use the FCC tensor instead of the original global state as input to the CVAE encoder to achieve a more practical and stable training process. The following procedures show how our designed FCC encoder works:

We first define $AT_{i}^{j}$ as the “expected shortest \textbf{A}rriving \textbf{T}ime” that the EV agent $i$ will spend to arrive at the EVCS $j$. $AT_{i}^{j}$ is calculated as :
\begin{align*}
AT_{i}^{j}=\sum_{edge \in L_{j}}d_{edge}/v_{edge} \tag{8}
\end{align*}
following the Dijkstra shortest path method based on the current traffic flow. Then we denote $CT_{i}^{j}$ as the “expected \textbf{C}harging \textbf{T}ime” that the EV agent $i$ will spend to charge at the EVCS $j$ after arrival. $CT_{i}^{j}$ is calculated as:
\begin{align*}
CT_{i}^{j} = (SOC_{i}^{max}-SOC_{i}^{arrive})/pow_{j}^{arrive} \tag{9}  
\end{align*}
based on the SOC of EV agent $i$ and charging power of EVCS $j$ when it arrives, where $pow_{j}^{arrive}$ represents the charging power of EVCS $j$ when EV $i$ arrives. We introduce an FCC tensor $fcc_i$ for each EV agent $i$, whose $j$ th element $fcc_i^j$ indicates the expected queueing time if EV $i$ selects EVCS $j$ as the charging destination. To calculate $fcc_{i}^{j}$, We focus on those EV agents that select the same spot at EVCS $j$ for charging and arrive earlier than EV agent $i$. There are three cases at that charging spot:

1) If no other EV agent arrives earlier than EV agent $i$, then the EV agent $i$ can directly charge at this spot and doesn't need to wait. In this case $fcc_{i}^{j}=0$.

2) If there is only one other EV agent $1$ arrives earlier than EV agent $i$, then EV agent $i$ has to wait until EV agent $1$ finishes charging. The remaining charging time of EV agent $1$ is $CT_{1}^{j} - (AT_{i}^{j} - AT_{1}^{j})$, thus $fcc_{i}^{j}=CT_{1}^{j} - (AT_{i}^{j} - AT_{1}^{j})$.

3) If there are $m$ EV agents ($m > 1$) already queuing at this spot, then EV agent $i$ has to wait until all of them finish charging. Thus $fcc_{i}^{j}=CT_{1}^{j} - (AT_{i}^{j} - AT_{1}^{j}) + \sum_{2}^{m} 
CT_{m}^{j}$. 

In conclusion, $fcc_{i}^{j}$ is calculated following equation (10):

\begin{equation}
fcc_{i}^{j} = 
\begin{cases} 
CT_{1}^{j} - (AT_{i}^{j} - AT_{1}^{j}) + \sum_{2}^{m} 
CT_{m}^{j}, & m \ge 2 \\ 
CT_{1}^{j} - (AT_{i}^{j} - AT_{1}^{j}), & m=1 \\
0 , & m=0
\end{cases} \tag{10}
\end{equation}

If there are $K$ EVCSs in this region, at each time step, a decision-making EV $i$ will maintain a fixed K-dimension FCC tensor as: 
\begin{equation}
\begin{aligned}
   FCC_{i} = softmax([fcc_{i}^{1}, … , fcc_{i}^{K}]) 
\end{aligned} \tag{11}
\end{equation}

The advantage of our proposed FCC-based encoder is shown in Fig. 8. As long as other EVs are expected to arrive later than EV $i$ at EVCS $j$ as destination, the FCC score $fcc_{i}^{j}$ are always $0$ regardless of the numbers and exact positions of other EVs. Through our FCC encoder, the large-scale global state can be simply represented by a value of 0 at the $j$-th element in EV $i$'s FCC tensor. Therefore we significantly simplify the high and varying dimensional input space into a lower and fixed dimensional FCC tensor, thus providing a more practicable and stable training of the model.

\begin{figure}[!ht]
\centering
\includegraphics[width=3in]{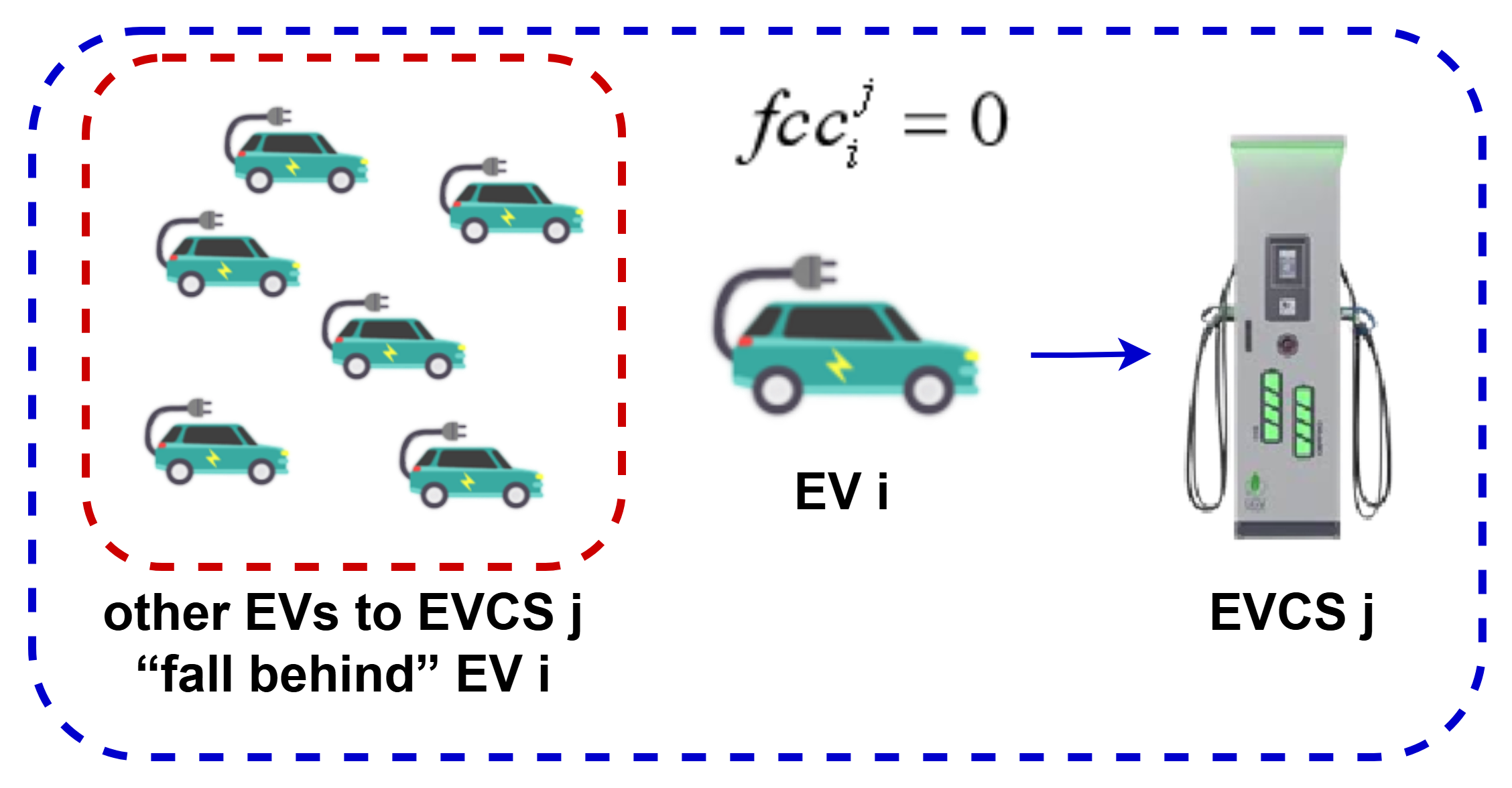}
\captionsetup{font=small}
\caption{An example that shows the advantage of our proposed FCC encoder. If other EVs “fall behind” EV $i$ when they are going to EVCS $j$, $fcc_{i}^{j}=0$ in all these cases regardless of the other EVs’ numbers and exact positions.}
\label{fig_sim}
\end{figure}

\subsubsection{Multi-gradient Descent Algorithm for Loss-balancing }

As shown in Fig. 4, we design a more practical framework where the EV agents make their own decisions partially based on the RI from the recommendation platform. However, the training of our generative model enhanced DRL method poses another great challenge: As the DQN network selects actions through the argmax policy, it doesn't require exact evaluation of each state-action pair's Q-value. Thus the DQN model is not sensitive to the fluctuation of the training loss. On the other hand, as the fixed-dimensional FCC tensor is compressed from the high-dimensional global state, it is extremely sensitive to the fluctuation of the loss as a subtle reconstruction bias of the FCC tensor may lead to a totally different global state. We observe that simply adding up the loss of the DRL network and CVAE model during training can cause severe instability of the performance of the algorithm. 

To address this challenge, we incorporate the Multi-gradient Descent Algorithm (MGDA) \cite{29} to balance the two parts of the loss functions, ensuring that each step of the gradient descent optimizes both components effectively. The loss-balancing problem between the DQN loss and the CVAE loss is formulated as the following optimization problem:

\begin{equation*}
\min_{\alpha \in [0, 1]} \left\| \alpha \nabla_{\theta^{sh}} \hat{\mathcal{L}}^{D}(\theta^{sh}, \theta^D) + (1 - \alpha) \nabla_{\theta^{sh}} \hat{\mathcal{L}}^{C}(\theta^{sh}, \theta^C) \right\|_2^2
. \tag{12}
\end{equation*}

\noindent where $\hat{\mathcal{L}}^{D}$ and $\hat{\mathcal{L}}^{C}$ denote the loss function of the DQN and CVAE model, respectively. $\theta^D$ and $\theta^C$ correspond to the parameters of the DQN and CVAE network. $\theta^{sh}$ denotes the shared parameters that the gradient pass through both the DQN and CVAE model. The objective of equation (12) is to minimize the L2-Norm square of the weighted sum of the gradients of the two loss functions. By adaptively adjusting the weight parameter $\alpha$, MGDA ensures that both of the loss decrease during each gradient descent step. The solution of $\alpha$ in (12) is:

\begin{equation*}
\centering
    \left[ \frac{(\nabla_{\theta^{sh}} \hat{\mathcal{L}}^{C}(\theta^{sh}, \theta^{C}) - \nabla_{\theta^{sh}} \hat{\mathcal{L}}^{D}(\theta^{sh}, \theta^{D}))^\top \nabla_{\theta^{sh}} \hat{\mathcal{L}}^{C}(\theta^{sh}, \theta^{C})}{\|\nabla_{\theta^{sh}} \hat{\mathcal{L}}^{D}(\theta^{sh}, \theta^{D}) - \nabla_{\theta^{sh}} \hat{\mathcal{L}}^{C}(\theta^{sh}, \theta^{C})\|_2^2} \right]_{+} 
    \tag{13}
\end{equation*}
where $[ \cdot ]_{+}$ represents clipping the element to $[0, 1]$ as $[a]_{+}=\max(\min(a, 1), 0)$. Thus, after gaining the gradients of $\theta^D$, $\theta^C$ and $\theta^{sh}$, following (13) we reweight each part at each decision step. The algorithm of MGDA is shown as follows:

\begin{algorithm}[!ht]
\caption{Adaptive weight balance of gradients $\theta^D, \theta^C$}
\begin{algorithmic}[1]
\IF{$(\theta^D)^T \theta^C \geq (\theta^C)^T \theta^D$}
    \STATE $\alpha = 1$
\ELSIF{$(\theta^D)^T \theta^C \geq (\theta^C)^T \theta^C$}
    \STATE $\alpha = 0$
\ELSE
    \STATE $\alpha = \frac{(\theta^C - \theta^D)^T \theta^C}{\|\theta^C - \theta^D\|_2^2}$
\ENDIF
\end{algorithmic}
\end{algorithm}

\section{CASE STUDIES}

\subsection{Experimental Setup}
To assess the effectiveness of the proposed approach, our algorithm is tested in a simulation based on a real zone within Xi'an City, mirroring the setup detailed in \cite{19}. The experiment is conducted on a Linux server with one NVIDIA RTX A5000 GPU serving as the experimental infrastructure. In terms of software, we utilize the PyTorch deep learning framework with a Python 3.8 environment.

Following the original configuration outlined in \cite{19}, we categorize the 39-nodes traffic roads into three classes, visualized with distinct colors: green denoting the ring highway encircling the city, yellow symbolizing the urban expressways, and red indicating the inner ring roads. The formulated 39-nodes traffic graph network is shown in Fig 9.

\begin{figure}[!ht]
\centering
\includegraphics[width=3in]{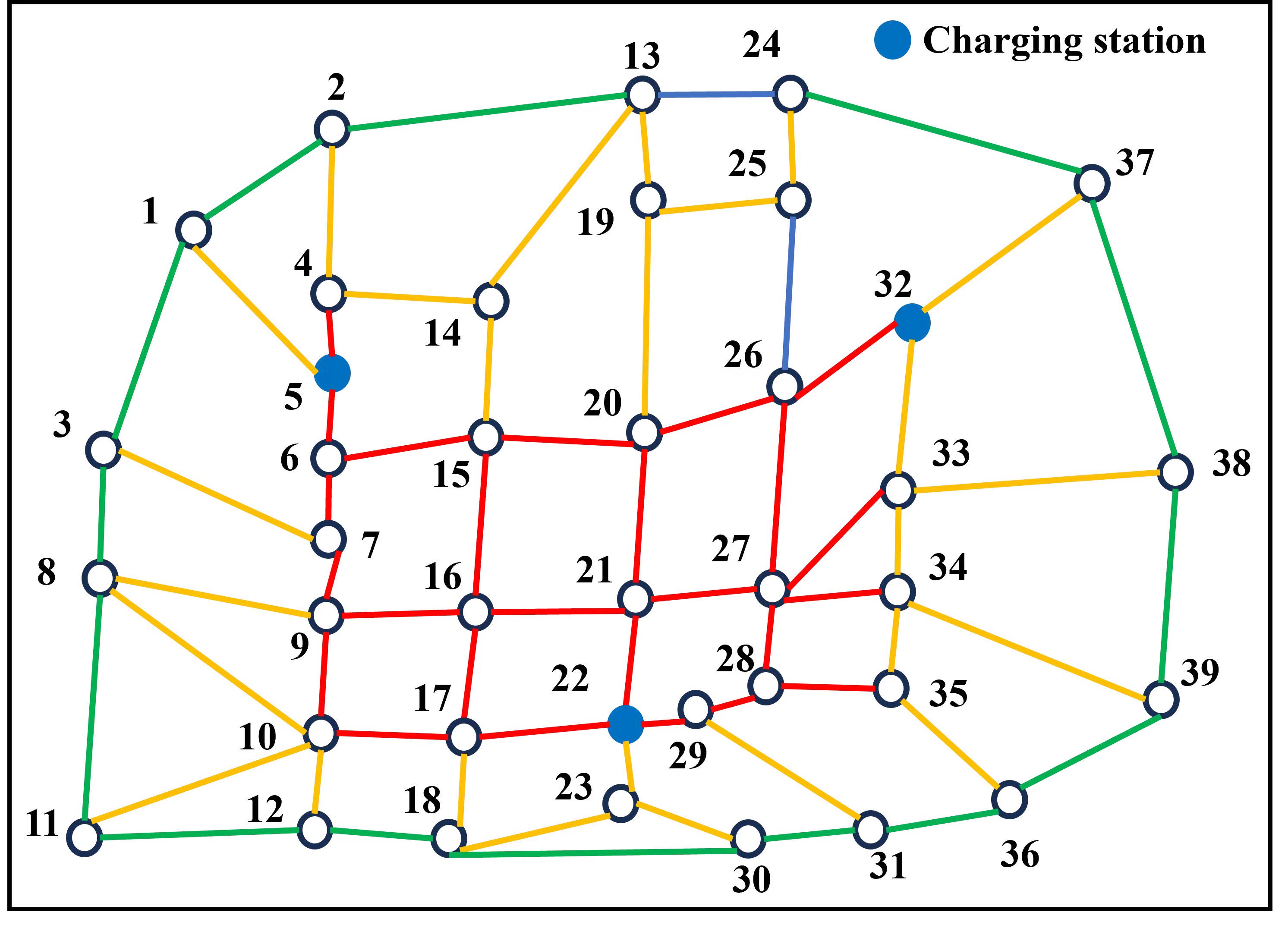}
\captionsetup{font=small}
\caption{Traffic graph of the simulation environment.}
\label{fig_sim}
\end{figure}

The settings of the random variables are outlined in Table II. EVs are required to abide by the specific speed limits corresponding to the different road groups. The original speed limits are set as 120 km/h, 80 km/h, and 60 km/h, respectively. Consequently, we establish the speed distributions with expected speeds of 90 km/h, 70 km/h, and 50 km/h based on these limitations. Furthermore, with the integration of renewable energy sources, electric charging prices may vary over time. Hence, we introduce an additional distribution for electric prices with an expected value of 0.45 yuan/kWh. To standardize the final cost, we convert the waiting time cost at charging stations into monetary units. Accordingly, we assign a value of 2 yuan per 5 minutes, aligning with the average waiting price for taxis in China during 2023 \cite{31}. The electricity price fluctuates every 30 minutes, while the traffic flow conditions change every 5 minutes. By navigating through these dynamic conditions, the EVs aim to optimize their routes and charging decisions to minimize the total costs.

\begin{table}[h]
  \centering
  \captionsetup{font=small}
  \caption{Random Variables in EV Charging Navigation}
  \begin{tabular}{p{2.5cm} p{2.5cm} c} 
    \toprule
    Random Variables & Distribution & Boundary \\
    \midrule
   Velocity on green roads (km/h) & $v_{\text{green}} \sim N(0.9 \ast 120, (0.05 \ast 120)^2)$ & $0 < v_{\text{green}} \leq 120$ \\ 
Velocity on yellow roads (km/h) & $v_{\text{yellow}} \sim N(0.7 \ast 80, (0.10 \ast 80)^2)$ & $0 < v_{\text{yellow}} \leq 80$ \\ 
Velocity on red roads (km/h) & $v_{\text{red}} \sim N(0.5 \ast 60, (0.15 \ast 60)^2)$ & $0 < v_{\text{red}} \leq 60$ \\ 
EV Charging price (\$/kWh) & $\lambda_k^{\text{ch}} \sim N(a, (0.15a)^2)$ & $0.3 \leq a \leq 0.7$ \\ 
EV initial SOC & $e^{\text{ini}} \sim U(0.4, 0.6)$ & $0.4 \leq e^{\text{ini}} \leq 0.6$ \\ 
    \bottomrule
  \end{tabular}
\end{table}

\subsection{Training Process}

In our algorithm, the training process involves two main components: the training of the DQN model and the CVAE-LSTM model. Both segments require several hyperparameters, including the learning rate, batch size, number of hidden layers, number of neurons, and discount factor. The specific hyperparameter settings are detailed in Table III.

\begin{table}[!ht]
  \centering
  \captionsetup{font=small}
  \caption{Hyperparameter setting of each model}
  \begin{tabular}{cc}  
    \toprule
    Hyperparameter &  Value \\
    \midrule
    learning rate of DQN & 0.0005 \\ 
    learning rate of CVAE & 0.00001 \\ 
    batch size & 16 \\ 
    discount factor & 0.99 \\ 
    hidden layers of LSTM & 2 \\
    layers of DQN & 3 \\
    layers of CVAE Encoder & 2 \\
    layers of CVAE Dncoder & 2 \\
    episodes & 1000 \\
    optimizer & ADAM \\
    replay buffer size & $10^{6}$ \\
    \bottomrule
  \end{tabular}
\end{table}

\subsection{Main Result}

We evaluate the effectiveness of our generative enhanced MARL algorithm by conducting comparative analyses against various baseline methods. These baselines include Independent Q-learning (IQL) \cite{19}, IQL with global information which is simply compressed by FCC encoder (IQL-global-FCC, serving as an ideal case upper bound), and another GNN-based DQN algorithm also utilizing global information called $BGRL$ \cite{24}. We also implement two CTDE DRL baseline algorithms: Muti-agent Actor-Critic (MA-AC) \cite{32} and Qmix \cite{33}. The performance assessments are conducted across the scenarios including 2 EVs and 20 EVs, providing an insight into the model's scalability and robustness across different scales.

\begin{figure}
    \centering
    \begin{subfigure}{0.45\textwidth}
        \centering
        \includegraphics[width=\textwidth]{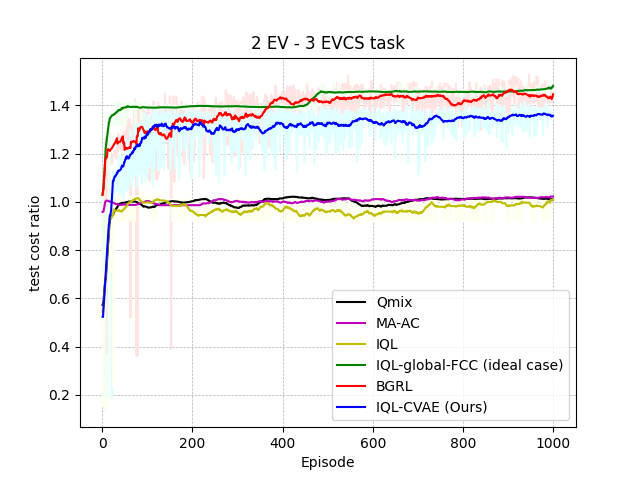}
        \captionsetup{font=small}
        \caption{2 EV scene result}
    \end{subfigure}
    \hfill
    \begin{subfigure}{0.45\textwidth}
        \centering
        \includegraphics[width=\textwidth]{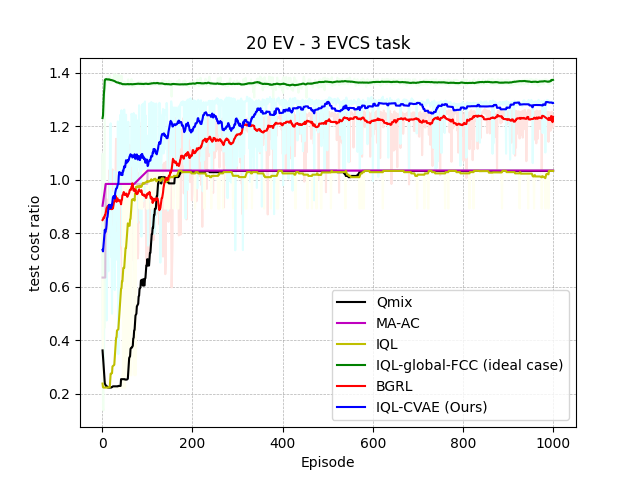}
        \captionsetup{font=small}
        \caption{20 EV scene result}
    \end{subfigure}
    \captionsetup{font=small}
    \caption{Main result of proposed navigation algorithm}
\end{figure}

Fig. 10 (a), (b) illustrate the performance of our algorithm. We evaluated the performance of each method by computing the “cost ratio”, which is calculated by dividing the overall cost of the Dijkstra shortest-path algorithm by the cost of each respective algorithm. A higher ratio indicates a lower total cost of that algorithm, which is a better method performance. 

The results for both scenarios are shown in Table IV, highlighting the significant performance enhancements achieved by our algorithm compared to fully decentralized execution method such as IQL, MA-AC and Qmix. Specifically, in the contexts of the 2 EV and 20 EV scenarios, our model demonstrates performance improvements exceeding 36\% and 27\% respectively over them. 

It is noteworthy to contrast these results with those algorithms that have unrestricted access to real-time information from all other EV drivers, as seen in IQL-global-FCC and BGRL. In this comparison, our method only exhibits a marginal performance drop of approximately 10\% in both scenarios compared with IQL-global-FCC. Particularly within the 20 EV scenario, our approach even surpasses BGRL's performance despite operating without global information access. 

\begin{table}[h]
\centering
\caption{Total cost ratio}
\begin{tabular}{ ccc}
\toprule
\textbf{Algorithm / the number of EVs} & \textbf{2 EV scene} & \textbf{20 EV scene}  \\

\midrule
\textbf{IQL-global-FCC (upper bound)}  & 1.44 $\pm$ 0.01 & 1.37 $\pm$ 0.01 \\
\textbf{IQL}  & 1.01 $\pm$ 0.03  & 1.02 $\pm$ 0.07 \\
\textbf{Qmix}  & 1.02 $\pm$ 0.03  & 1.03 $\pm$ 0.01 \\
\textbf{MA-AC}  & 1.02 $\pm$ 0.01  & 1.03 $\pm$ 0.01 \\
\textbf{BGRL}  & \textbf{1.42 $\pm$ 0.06} & 1.24 $\pm$ 0.21 \\
\textbf{IQL-CVAE (Ours)} & 1.36 $\pm$ 0.08 & \textbf{1.27 $\pm$ 0.22} \\

\bottomrule
\end{tabular}
\end{table}

An important observation we have noted is that compressing global data by our designed FCC encoder (IQL-global-FCC) instead of with GNN (BGRL) significantly accelerates the training process and enhances the stability and performance. Notably, as the number of EVs increases, the dimension of the input space expands drastically. In BGRL, despite the utilization of the graph attention to embed global information, it still maintains a one-to-one mapping from its original global state to an embedding latent input. As discussed earlier, the core issue introduced by partial observations is the FCC problem, suggesting that dynamic global states may lead to similar queuing times at the EVCS due to the same FCC impact of them. Consequently, our designed FCC encoder adeptly captures the vital information latent in the original data, establishing a multi-to-one mapping from the global state to the fixed-dimensional FCC tensor. These advantages underscore the efficacy of our proposed rule-based encoder in succinctly condensing the state space and streamlining the training process.

\subsection{Ablation Study}

\begin{figure}
    \centering
    \begin{subfigure}{0.45\textwidth}
        \centering
        \includegraphics[width=\textwidth]{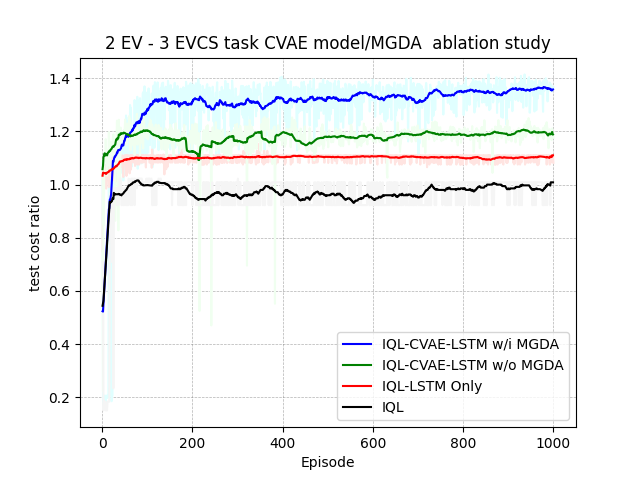}
        \captionsetup{font=small}
        \caption{2 EV scene ablation result}
    \end{subfigure}
    \hfill
    \begin{subfigure}{0.45\textwidth}
        \centering
        \includegraphics[width=\textwidth]{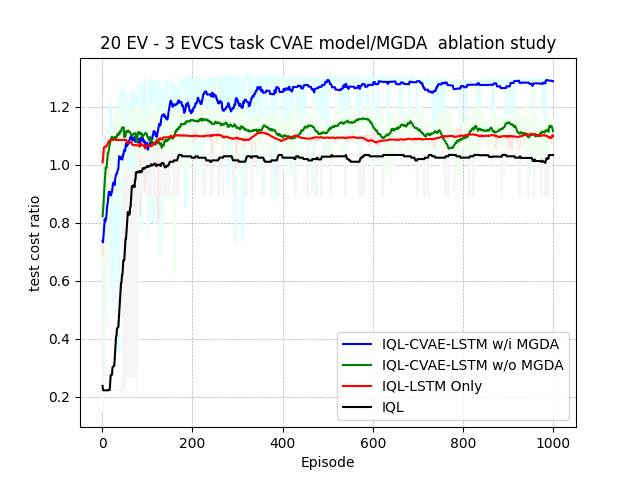}
        \captionsetup{font=small}
        \caption{20 EV ablation result}
    \end{subfigure}
    \captionsetup{font=small}
    \caption{CVAE / MGDA ablation study}
\end{figure}

In Fig. 11 (a), (b), we compare the methods with and without CVAE modules, and the implementation of MGDA. The exact results are shown in Table V. We observe that introducing only the LSTM encoder's output as RI without CVAE model to feed into DQN model, yields a modest performance improvement of merely 11\% and 8\% over the shortest-path algorithm in the  scenario of 2 EV and 20 EV. With MGDA, the performance of the algorithm reaches 16\% and 14\% better, compared with the algorithm without MGDA. These results show the significance of integrating the CVAE model and MGDA, enabling EV agents to get a higher performance gain.

\begin{table}[h]
\centering
\captionsetup{font=small}
\caption{Total cost ratio of ablation result}
\begin{tabular}{ ccc}
\toprule
\textbf{Algorithm / the number of EVs} & \textbf{2 EV scene} & \textbf{20 EV scene}  \\

\midrule
\textbf{IQL-LSTM-CVAE w/i MGDA} & \textbf{1.36 $\pm$ 0.08} & \textbf{1.27 $\pm$ 0.22} \\
\textbf{IQL-LSTM-CVAE w/o MGDA} & 1.20 $\pm$ 0.05 & 1.13 $\pm$ 0.18 \\
\textbf{IQL-LSTM encoder only}  & 1.11 $\pm$ 0.03  & 1.08 $\pm$ 0.11 \\
\textbf{IQL}  & 1.01 $\pm$ 0.03  & 1.02 $\pm$ 0.07 \\

\bottomrule
\end{tabular}
\end{table}

\begin{figure}[!ht]
\centering
\includegraphics[width=3in]{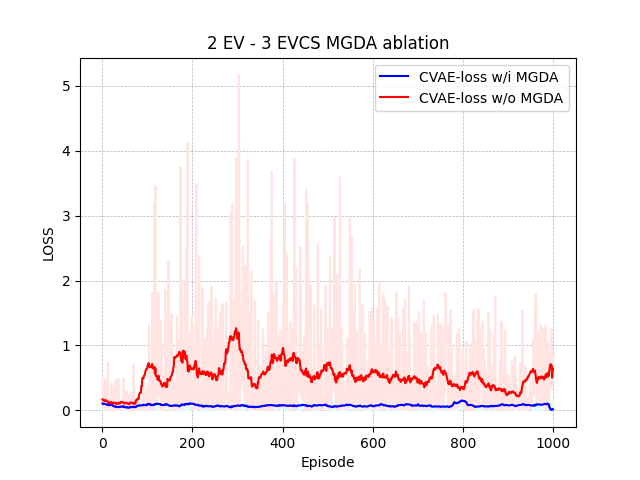}
\captionsetup{font=small}
\caption{MGDA for degradation of CVAE loss.}
\label{fig_sim}
\end{figure}

Fig. 12 further validates the specific impact of the MGDA to balance the gradient descent steps of both the DQN and CVAE models. As illustrated in section III, the CVAE model is highly sensitive to the fluctuation of loss. Even slight biases in RI can lead to significant errors in reference to original global state. As shown in Fig. 12, the CVAE loss remains at 0.49 after 1000 training episodes by simply adding the DQN and CVAE losses together. But the CVAE loss significantly decreases to 0.08 after introducing MGDA, emphasizing the necessity of the MGDA method to dynamically balance the updated steps between the DQN and CVAE models.

\section{Conclusion}

This paper introduces a novel generative model enhanced multi-agent DRL algorithm for the real-time EV charging navigation task. We integrate a CVAE model with an LSTM encoder to provide RI solely based on EV agent's local observation. Besides, we introduce a novel FCC-based encoder, which effectively compresses the global state into a fixed-dimensional FCC tensor to provide a more stable foundation during the training process. Additionally, we employ MGDA to dynamically balance the weights between the gradients of the DQN loss and the CVAE loss, leading to enhanced training stability. We develop a simulator based on part of real city map to conduct experiments. The simulation results demonstrate that our method achieves comparable performance with only about an 8\% decrease compared to methods that have access to global information. 

\bibliographystyle{IEEEtran}
\bibliography{IEEEabrv,Bibliography}

\end{document}